\documentclass[letterpaper, 10 pt, conference]{ieeeconf}
\IEEEoverridecommandlockouts
\usepackage{times}
\usepackage{soul}
\usepackage{url}
\usepackage[hidelinks]{hyperref}
\usepackage[utf8]{inputenc}
\usepackage[small]{caption}
\usepackage{graphicx}
\usepackage{amsmath}
\usepackage{booktabs}
\usepackage{algorithm}
\usepackage{algorithmic}
\usepackage{amsmath}
\usepackage{latexsym}
\usepackage{booktabs}
\usepackage{subfig}
\usepackage{array}
\usepackage{multirow}
\usepackage{cite}
\newcolumntype{L}[1]{>{\raggedright\let\newline\\\arraybackslash\hspace{0pt}}m{#1}}
\newcolumntype{C}[1]{>{\centering\let\newline\\\arraybackslashspace{0pt}}m{#1}}
\newcolumntype{R}[1]{>{\raggedleft\let\newline\\\arraybackslash\hspace{0pt}}m{#1}}
\usepackage{lipsum}

\renewcommand\footnotemark{}
\title{
A LiDAR Assisted Control Module with High Precision in Parking Scenarios for Autonomous Driving Vehicle
}

\author{
	Xin Xu$^{1}$,
	Yu Dong$^{1}$,
	Fan Zhu$^{2, *}$ \\
    \thanks{$^{*}$Contact Author.}
    \thanks{$^1$ Baidu Inc., Haidian District, Beijing, China.}
    \thanks {\{xinme\}@live.com; \{dongyu02\}@baidu.com}
    \thanks{$^2$ Baidu USA LLC, Sunnyvale, CA 94089, USA.}
    \thanks{\{fanzhu\}@baidu.com}
}

\begin{document}

\maketitle
\thispagestyle{empty}
\pagestyle{empty}
\begin{abstract}
Autonomous driving has been quite promising in recent years. The public has seen Robotaxi delivered by Waymo, Baidu, Cruise, and so on. While autonomous driving vehicles certainly have a bright future, we have to admit that it is still a long way to go for products such as Robotaxi. 
On the other hand, in less complex scenarios autonomous driving may have the potentiality to reliably outperform humans. For example, humans are good at interactive tasks (while autonomous driving systems usually do not), but we are often incompetent for tasks with strict precision demands. 
In this paper, we introduce a real-world, industrial scenario of which human drivers are not capable. The task required the ego vehicle to keep a stationary lateral distance (\textit{i.e.} 3$\sigma$ $\leq$ 5 centimeters) with respect to a reference. To address this challenge, we redesigned the control module from Baidu Apollo open-source autonomous driving system. 
A precise (3$\sigma$ $\leq$ 2 centimeters) Error Feedback System was first built to partly replace the localization module. Then we investigated the control module thoroughly and added a real-time calibration algorithm to gain extra precision. We also built a simulation to fine-tune the control parameters. 
After all those works, the results are encouraging, showing that an end-to-end lateral precision with 3$\sigma$ $\leq$ 5 centimeters has been achieved. Further, we show that the results not only outperformed original Apollo modules but also beat specially trained and highly experienced human test drivers.

\textbf{Keywords:} Autonomous Driving, High Precision Parking, Localization, Control.
\end{abstract}

\section{Introduction}
In the past a few years, autonomous driving has been intensively studied and discussed. 
The community has seen tremendous progress made on perception \cite{Chen2015DeepDriving, Casser2019Unsupervised, Gordon2019Depth, Sun2019Scalability, 2020PointContrast}, prediction \cite{Alahi2016Social, Vemula2018Social, Deo2018How, Pan2019Lane, Xu2020Data}, simulation \cite{Benekohal1988CARSIM, Dong2011Driver, Geoffrey2011Driving, CARLA2017}, etc. 
Interestingly, there are relatively less literature focused on control in autonomous driving, although it is a very mature topic developed over one hundred years. One reason is that control is usually designed to track planned trajectory, and unfortunately there are plenty unsolved problems in planning \cite{RN290}. That said, control module can indeed contribute to autonomous driving on its own. In 2020, we have shown that a control module with longitudinal calibration algorithm improves tracking ability considerably \cite{9304778}. In this paper, we introduce a redesigned control module with improved lateral control algorithms at a level of lateral precision down to 5 centimeters (cm). That is, the ego vehicle is able to keep a lateral distance within $\pm$ 5cm, with respect to a reference.

Recently, as we simultaneously work with various OEMs across a large range of topics, we have noticed that, in certain scenarios (such as parking), localization and control performance are the key to the success. 
For example, for a vehicle that serves for both ordinary and disabled people, the ability of the vehicle to stop precisely along the curb, \textit{i.e.} target $\pm$ 5cm, shows whether the autonomous vehicle is more friendly than a traditional one or not --- since disabled people with a wheelchair can simply steer it to get on the bus without extra help if the gap between the vehicle and the curb is small enough. 
Moreover, a truck that works in an automated port may need to park in a specific area with a precision of few centimeters, to avoid failure to load/unload containers. 
In those scenarios, one can expect human drivers, even those well-trained, certified professional test drivers, to perform incompetently. 
We were then motivated to provide a solution based on an open-source autonomous driving platform (Baidu Apollo) for those scenarios, not only because they are good showcases for autonomous driving, but also they are real industrial demands.

Generally, a system's precision depends on various factors, such as localization, HD (High-Definition) map, control, sensor, actuator, system delay, and even weather and road surface, not to mention that different factors often interact. Hence, it would be very difficult to inspect all factors individually and thoroughly. 
To address this issue, in this paper, we roughly split factors into two groups, namely controllable factors and uncontrollable factors. 
Controllable factors mainly include software parts, i.e, localization, HD map, control, system delay, whereas uncontrollable factors include factors such as actuator, sensor, weather, and road condition. Of course, with more resources one could transfer uncontrollable factors into controllable ones, for example, to build a new actuator and/or a new sensor. 
Nevertheless, we aimed to provide a solution that best suits most autonomous platforms with an affordable cost and minimum modifications. As a result, this paper will focus on software part, and we will show that the lateral precision was indeed improved significantly with modified software, with other conditions remained the same (same sensor, same vehicle, same weather condition, same road, etc.). Further, we divided controllable parts into external and internal factors from a control module's perspective. External factors are signals sent to control module, such as localization, HD map, and system delay. Internal factors are signals processed within control module, such as steering-wheel offset (a vehicle's intrinsic property), heading offset caused by IMU (Inertial measurement unit) mounting error. Based on this concept, this paper will show how we addressed external and internal factors individually and integrated them to eventually achieve an extra high lateral precision.

Taken altogether, in this paper we present an algorithmic architecture that integrates existing work with Baidu Apollo autonomous driving system \cite{Apollo2017, Fan2018} to solve a real-world problem. Yet, we show that the results were far better than human drivers, bringing the community an example that autonomous driving system outperforms human drivers in real-world, industrial scenarios. 
This paper is organized in the following way: i). Section \ref{sec:method} introduces the methodology and compare it with existing ones; ii). Section \ref{sec:result} presents experimental set-up and shows the results; Section \ref{sec:conclusion} concludes the work. 


\section{Method} \label{sec:method}

\begin{figure}[thpb!]
\vspace{-0.cm}
\small
\centering
    \subfloat{\includegraphics[width= 1.0\linewidth]{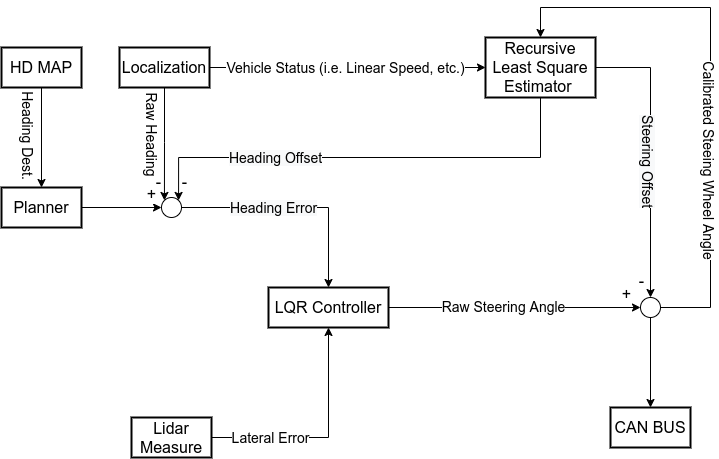}} \\    
  \caption{This figure shows the workflow of proposed method. Every Block represents a function module, 
whereas arrow lines connecting blocks represent signals.}
  \label{fig:workflow}
  \scriptsize{}
\end{figure}

\subsection{Workflow}
Figure \ref{fig:workflow} shows the architecture of this solution. The architecture can be elaborated from two aspects. First, the external factors, i.e., HD map, localization, and system delay. Fig. \ref{fig:workflow} shows that HD map only provides final heading state, i.e., the heading state the ego vehicle should achieve upon full stop, and localization only provides real time heading feedback. Neither the HD map tells the system where the reference is, nor the localization module tells the system how far away it is from that reference. Some may wonder why not build references into HD map and then use localization module in the most common way \cite{Durrant2006Simultaneous, Durrantwhyte2006Simultaneous}. The reasons are less intuitive, one can think of that during map collection there is only GNSS-based localization available, which is easily affected by the quality of GNSS signals and the distance between ego vehicle and base station. Furthermore, errors on LiDAR put another burden on map production in terms of accuracy and precision. Thus, HD map's precision and accuracy are affected by both localization and LiDAR performance, and localization performance (with HD map), in turn, is affected by map's accuracy and precision. This makes it extremely hard to analyze what the localization and HD map need to achieve to guarantee the lateral precision this study pursues. Another example can be seen in a paper published by Apollo localization team in 2018 \cite{8461224}, in which the team showed the best performance of the Apollo localization (with point-cloud HD map) was lateral RMS (root mean square error) around 4cm, with 3$\sigma$ around 30cm. Obviously, current localization and HD map technology are likely not fully ready for a system with precision within 3$\sigma$ $\leq$ 5cm. It is therefore that this paper only used HD map as a heading reference, instead of a reference with an absolute position. Correspondingly, localization module in this paper only measured ego vehicle's heading, instead of its absolute position. Notice that the system needs heading feedback as an input (see Fig. \ref{fig:workflow}), and we did not find a better of providing heading estimation other than localization module. Simultaneously, lateral error between ego vehicle and its targeting reference was estimated by LiDAR directly, which not only increased precision (only 1-2cm measurement error) but also reduced system delay (reduced around 100 milliseconds), see later sections for details. Reduction on time delay played an important role in maintaining precision, since a vehicle running at 10km/h can travel around 3cm every 100ms. As to the internal factors, i.e., the steering wheel offset and heading offset, this paper used RLS (recursive least square estimator) \cite{Narayanan2014Methods} to estimate them in a real-time fashion. It is worth mentioning that those offsets are often very small in practice and hence are neglected in most autonomous driving solutions. That said, for a system with as high as 5cm precision requirement, every offset plays a role in error contribution and therefore needs to be addressed seriously.

To summarize, in this workflow we first revised the external factors to provide more precise feedback, after which we calibrated internal signals to make the final output as correct as possible. Fig. \ref{fig:workflow} explains how we integrated all those efforts to generate higher quality of signal inputs (see heading error and lateral error), as well as correct output (see steering wheel angle output).

\subsection{Error Feedback System}
\begin{figure*}[thpb!]
\vspace{-0.cm}
\small
\centering
    \subfloat[Static]{\label{fig:static}\includegraphics[width= 0.48\linewidth]{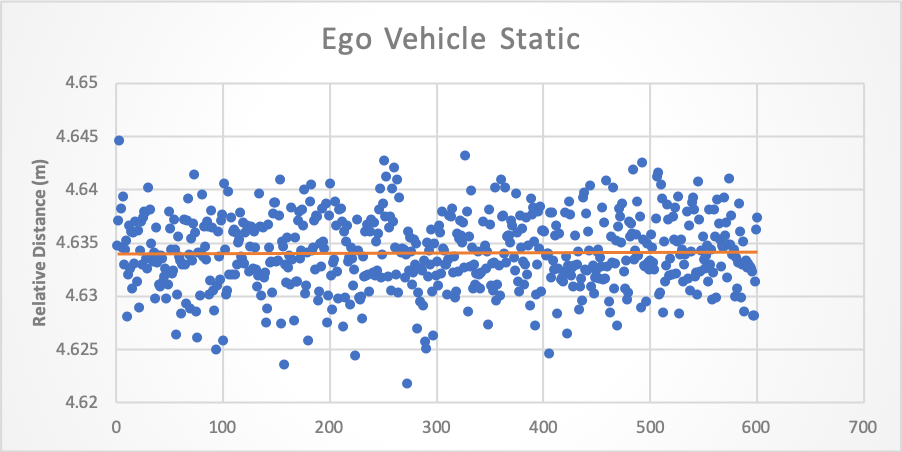}}
    \subfloat[Moving]{\label{fig:moving}\includegraphics[width= 0.48\linewidth]{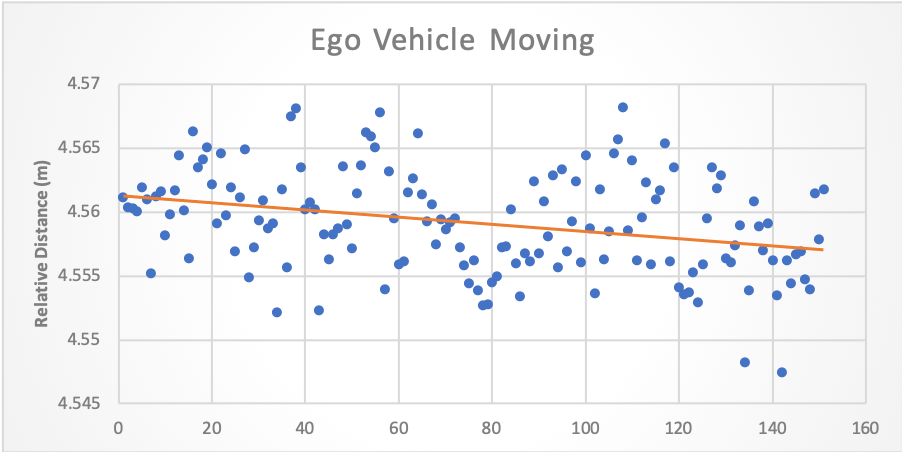}} \\    
  \caption{This figure illustrates distance measurement relative to a reference from LiDAR, with ego vehicle either static (left) or moving parallel to that reference (right). Note that data mainly clustered in a zone about 1cm wide (see vertical axis). The orange line represents the estimated line using Least Square. With the vehicle static, std. of the lateral error is 0.34 centimeters. With the vehicle moving (as parallel to the reference as possible and as less steering as possible, manual driving), std. of that is 0.42 centimeters}.
  \label{fig:distance_measure}
  \scriptsize{}
\end{figure*}
In order to achieve an end-to-end precision of 3$\sigma$ $\leq$ 5cm, the error feedback should be even more precise, \textit{e.g.} 3$\sigma$ $\leq$ 2cm.
In this paper, the key was to use LiDAR to do the measurement on lateral error. LiDAR is a piece of standard equipment used in almost all Level-4 autonomous driving vehicles for perception and localization \cite{Levinson2011Towards}. 
The LiDAR we used has the capacity of 1 to 2cm precision (according to Hesai Pander40P LiDAR specs \cite{Hesai40P}), way better than that of a typical localization and/or HD map solution \cite{2019L, Zhou2020DA4AD, 8461224}. The reason we did not use HD map based on this LiDAR is that errors in map production does not only come from LiDAR but also localization during data collection. That is, a LiDAR with 1 to 2cm precision leads to a HD map with larger error. 
On the other hand, heading feedback still came from localization and HD map modules, because a single LiDAR is simply not capable of providing heading estimation. One may wonder whether the relatively less precise heading estimation (from localization) would affect the overall performance. In fact, since we improved the lateral precision by an order of a magnitude (\textit{i.e.} from 1$\sigma$ $\leq$ 10cm in localization (with HD map) to 3$\sigma$ $\leq$ 2cm), the overall precision should benefit significantly just from this. Through direct LiDAR measurement, we also reduced system delay around 100ms ~ 160ms, since there was no other modules between LiDAR and control algorithm. By contrast, previously lateral error was successively processed by HD map, localization, and planner, resulting to a time delay when eventually passed to control module. Although a compensation to such delay is possible, error was still unnecessarily introduced.

Field tests on lateral error feedback from LiDAR, with ego vehicle either static or moving, proved the measurement was sufficiently precise (\textit{i.e.} 3$\sigma$ $\leq$ 2cm), see Figure \ref{fig:distance_measure}. We should mention that Fig. \ref{fig:distance_measure} only shows the standard deviation of this LiDAR was indeed around 1cm during either static or moving, it by no means verify the accuracy of its measurement. In practice, it would require a device with 10 times more precise than the origin one to verify the measurement of the origin one. In this case, if one doubts the ground truth of the lateral error measured by this LiDAR, one should acquire a device with 1 to 2mm precision to do the verification. In this paper, we took the official manual provided by Hesai as a guarantee and verified its standard deviation. 

\subsection{Control Algorithm, Modeling, and Simulation}
\begin{figure}[thpb!]
\vspace{-0.cm}
\small
\centering
    \subfloat{\includegraphics[width= 0.98\linewidth]{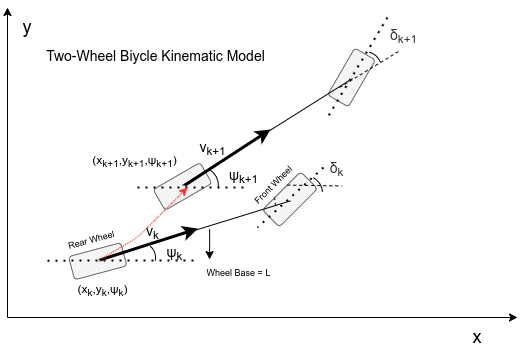}} \\    
  \caption{Kinematics model of the control algorithm}
  \label{fig:model}
  \scriptsize{}
\end{figure}
The control algorithm (Linear–Quadratic Regulator, \cite{LQR1974}) was designed under the assumption that ego vehicle often drives at a low to medium speed (\textit{i.e.} 0 to 40 KM/H). This assumption is valid due to that for parking scenarios the ego vehicle often spends a large amount of time cruising at low speed and eventually reach to a stop. Figure \ref{fig:model} illustrates the kinematics model under such assumption, which can be described mathematically as below,
\begin{equation}
\left\{
             \begin{array}{lr}
             x_{k+1} = x_{k} + v_k * cos(\psi_{k}) * \Delta t, &  \\
             y_{k+1} = y_{k} + v_k * sin(\psi_{k}) * \Delta t, & \\
             \psi_{k+1} = \psi_{k} + v_k * \frac{tan(\delta_k)}{L} * \Delta t &  
             \end{array}
\right.
\label{eq:model}
\end{equation}
where ($x$, $y$), $\psi$, $v$, $\delta$, and $L$ refers to the position, heading, linear speed, front-wheel angle, and the wheelbase of the ego vehicle, respectively.
Position and heading was obtained in a way described in the section of Error Feedback System (also see Figure \ref{fig:workflow}), while linear speed was provided directly by IMU. 
The front-wheel angle, on the other hand, maps to the steering wheel angle through a transfer function below.
\begin{equation}
    G(s) = \frac{1}{\tau s+1}
\end{equation}
The time constant $\tau$ in Eq. (2) was fine-tuned in simulation and field tests, and was set to 0.1668 as a result.

With this model, one can build a decent LQR controller but not the precise one as this paper purposed. The reason is that there are some subtle gaps between the model and the actual vehicle, although in most cases they are ignorable.
This model implies that one can get variables such as front-wheel angle and heading estimation ideally, which is incorrect due to errors such as steering-wheel offset and IMU mounting issue. Those errors are only negligible when they are small and extra precision is not desired. A steering-wheel offset may be caused by installation problem in factory, while heading offset in this case is from an askew mounted IMU. 
Figure 5 presents an example, showing how an askew mounted IMU influences the heading feedback. Note that position O represents the desired mounting point, while O' $(x_{offset}, y_{offset})$ represents the actual mounting point. $h_o$ is the resulting heading offset. 
\begin{figure}[thpb!]
\vspace{-0.cm}
\small
\centering
    \subfloat{\includegraphics[width= 0.65\linewidth]{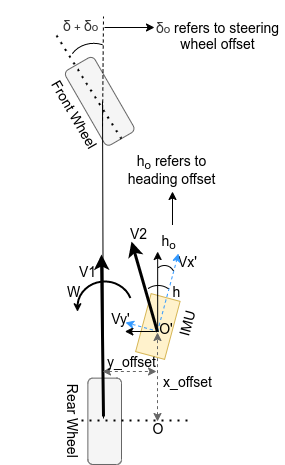}} \\    
  \caption{Heading offset due to incorrect mounting of IMU}
  \label{fig:heading_offset}
  \scriptsize{}
\end{figure}
In order to calibrate those offsets, the follow equations were deduced from the kinematic model (\ref{eq:model}) and Fig. \ref{fig:heading_offset}.
\begin{equation}
\left\{
             \begin{array}{lr}
             tan(\delta + \delta_{o}) = \frac{L\omega}{v_1}, & \\
             v_1 = v_2 * cos(h - h_o) - \omega * y_{offset} & \\
             v_2*sin(h-h_o) = \omega * x_{offset}, &  \\
             \end{array}
\right.
\label{eq:calibration}
\end{equation}
where $\delta_{o}$ represents the steering-wheel offset, and $h_o$, $y_{offset}$, and $x_{offset}$ represent the heading error, lateral error, and longitudinal error, respectively, due to incorrect mounting of IMU. 
It should be emphasized that the calculation of $v_1$ adds complication on offset estimation. In principle, $v_1$ is the linear speed of the ego vehicle while $v_2$ is the linear speed read from IMU. If the IMU is perfectly mounted or the vehicle goes perfectly straight (with no yaw rate at all), the two values match. In reality, however, $v_1$ is not observable and one can only estimate it from $v_2$. 
Now, both $h_o$ and $y_{offset}$ are involved in calculating $v_1$ from $v_2$, but we are only estimating $h_o$, which means we should acquire $y_{offset}$ from somewhere else. Of course, one solution is to estimate $y_{offset}$ as well, but it would make the estimation non-linear and hence difficult to solve online in a real-time manner. 
Therefore, we set $y_{offset}$ to 0.2 meter via carefully checking the mounting point and CAD (computer-aided design) model of the ego vehicle. Some may wonder whether a (reasonable) guess on $y_{offset}$, instead of a mathematical estimation, affects the precision or not. It can be proven that, for a typical mounting error with $h_o$ at 0.01 rad and $y_{offset}$ at 0.2m, the difference between $v_1$ and $v_2$ is less than $1\%$ on $\omega$ $\leq$ 0.05 rad/s and $v_2$ $\geq$ 1m/s, further leading to a calibration error on steering-wheel offset less than $0.01\%$. From this calculation, it is clear that even with $y_{offset}$ set to zero, the calibration error is negligible, not to mention a reasonable measurement on $y_{offset}$. Notice that it is not recommended to estimate $x_{offset}$ and/or $h_o$ from the CAD model too, because too many rough estimations rapidly increase the risk of breaking the system's
precision. We should always precisely estimate as many variables as possible. Further, it can be proven that a (un-calibrated) $h_o$ at 1 degree will lead to an lateral offset between front wheel and rear wheel around 7cm. One should always estimate $h_o$ as accurate as possible. As to estimate $h_o$ and $\delta_{o}$, Eq. \ref{eq:calibration} can be transformed to
\begin{equation}
\left\{
        \begin{array}{lr}
         \frac{L\omega}{v_1}-tan(\delta) = (1+\frac{L\omega}{v_1})tan(\delta_{o}), & \\
        atan(\frac{vy'}{vx'}) = \frac{\omega x_{offset}}{v_2} + h_o, & \\
        \end{array}
\right.
\end{equation}
where $v_x'$ and $v_y'$ refer to linear speed along x-axis and y-axis, respectively, of the IMU body frame. 
Obviously, this is a standard form of a least square problem:
\begin{equation}
    y = \phi\theta
\end{equation}
where 
\begin{equation}
\left\{
        \begin{array}{lr}
        y = [ \frac{L\omega}{v_1}-tan(\delta), atan(\frac{v_y'}{v_x'})]^T, & \\
        \theta = [tan(\delta_o), x_{offset}, h_o]^T, & \\
        \begin{split}
            \phi = \left[
                \begin{array}{ccc}
                 (1+\frac{L\omega}{v_1})& 0 & 0 \\
                  0 & \frac{\omega}{v_2} & 1 \\
                    \end{array}
                    \right]
            \end{split}
        \end{array}
\right.
\end{equation}

Directly, we can get estimations from the standard least square form, which is set to minimize the following loss function:
\begin{equation*}
    V(\hat{\theta}, n) = \frac{1}{2}\sum_{i=1}^{n}(y(i)-\phi^T(i)\hat{\theta})^2
\end{equation*}
Nevertheless, during driving we need to update estimations every frame as new data keep coming in. 
Fortunately, we can indeed use least square in a recursive form (Recursive Least Square, RLS for short). 
Through RLS, one can get estimations, \textit{i.e.} $\hat{\theta}$, in the following form:
\begin{equation}
    \hat{\theta}(k) = \hat{\theta}(k-1) + L(k)(y(k) - \phi^T(k)\hat{\theta}(k-1))
\end{equation}
where 
\begin{equation}
    L(k) = P(k-1)\phi(k)(1+\phi^T(k)P(k-1)\phi(k))^{-1}
\end{equation}
and
\begin{equation}
    P(k) = (1 - L(k)\phi^T(k))P(k-1)
\end{equation}
An initial value of $P(0)$ is needed to get RLS started.
In fact, $P(0)$ is related to the confidence of the initial guess of $\hat{\theta}$.
One can simply set $P(0)$ to a large value (such as 1e6) and initial guess of $\hat{\theta}$ to zero to get RLS started.
Figure 5 (a) and Figure 5 (b) show calibration results for steering and heading offset, respectively.
Figure 5 (c) presents a simulation result, in which one can see that the models built in this section (\textit{i.e.} simulated lateral/heading error) matches reasonably well with the actual vehicle data (\textit{i.e.} actual lateral/heading error).
\begin{figure}[thpb!]
\vspace{-0.cm}
\small
\centering
    \subfloat[Steering Wheel Angle Offset]{\label{fig:swa}\includegraphics[width=0.7\linewidth]{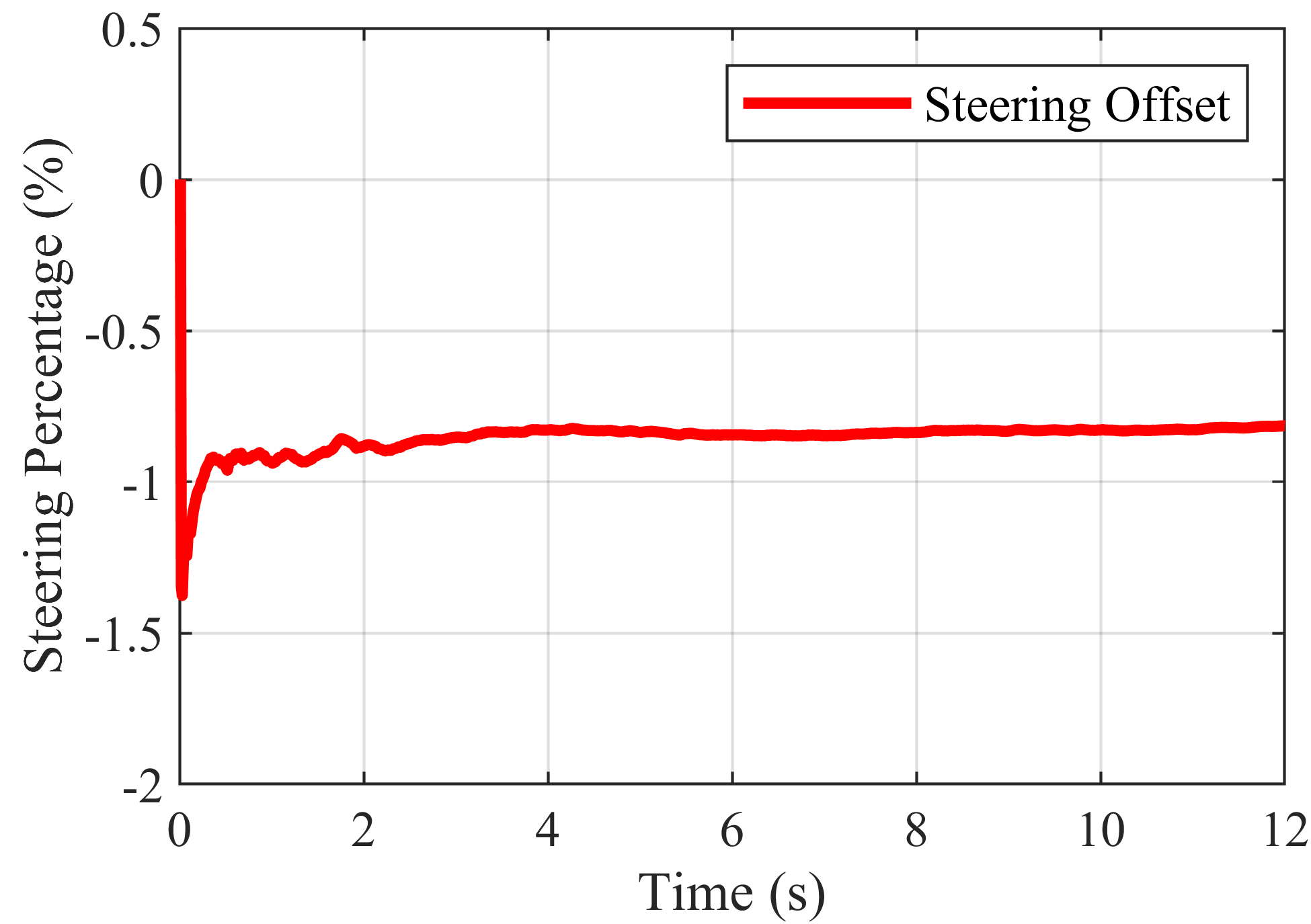}}\\
    \subfloat[IMU Heading and X offset]{\label{fig:imu}\includegraphics[width=0.7\linewidth]{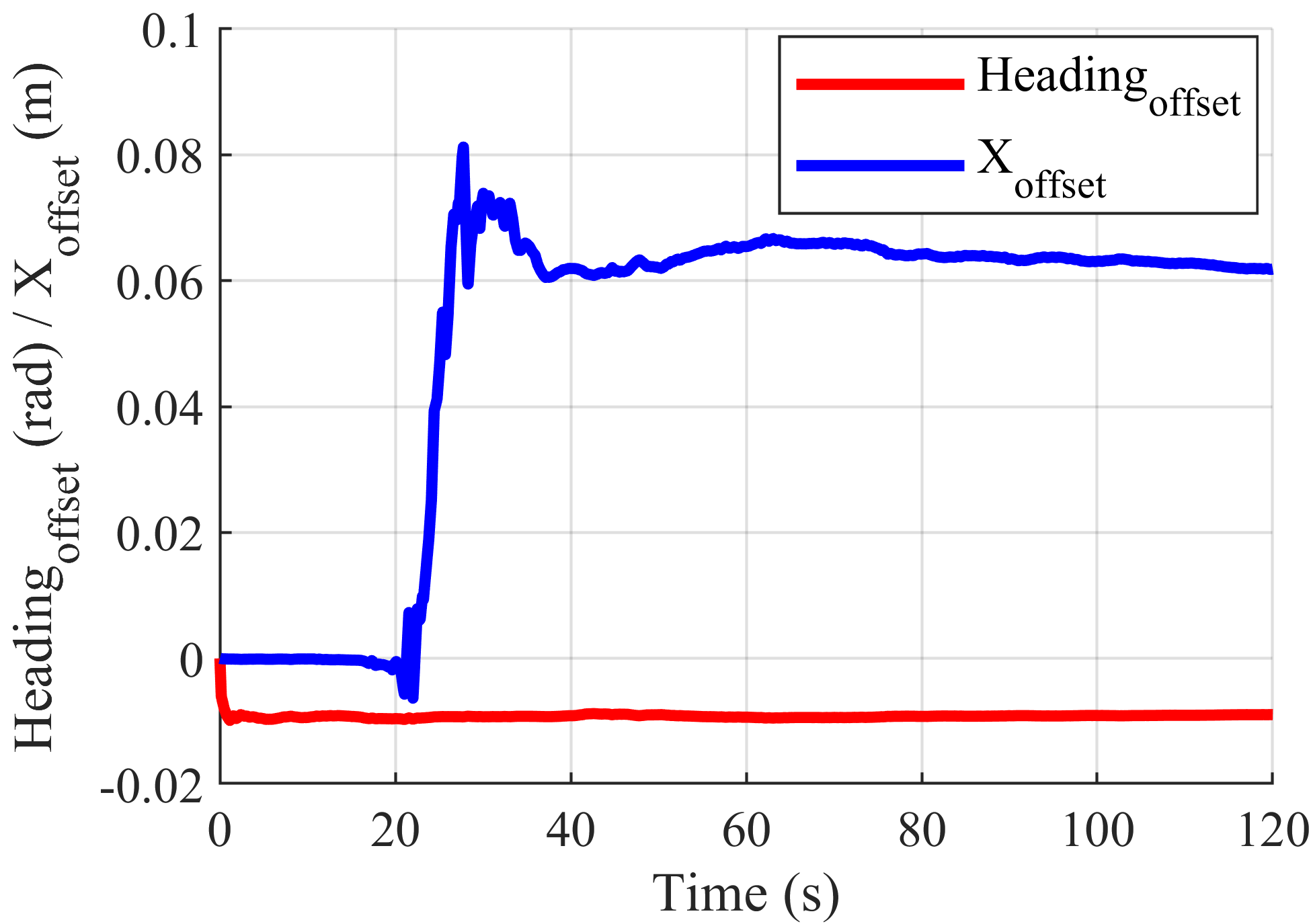}}  \\
    \subfloat[Simulated results and actual results]{\label{fig:imu}\includegraphics[width=0.7\linewidth]{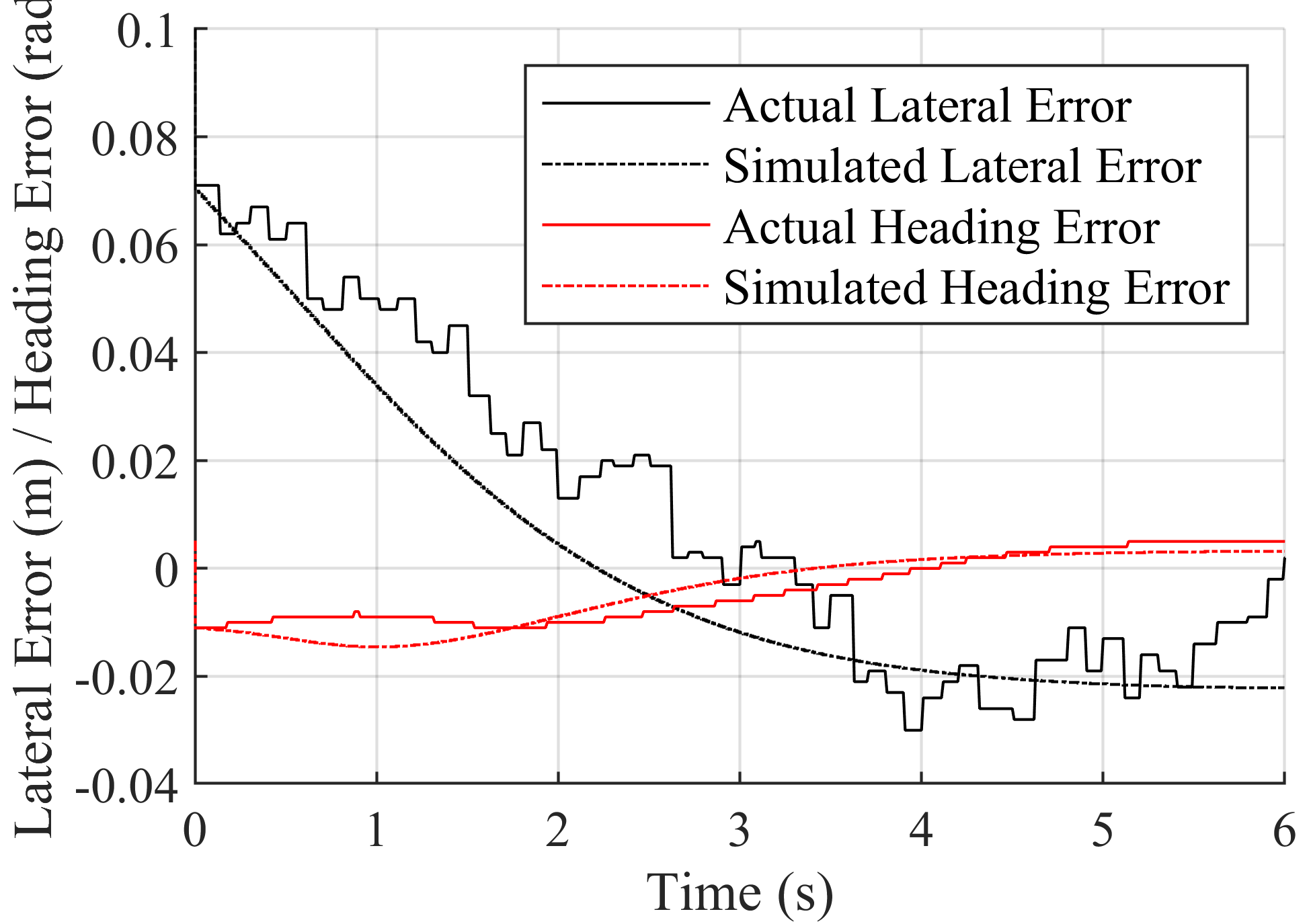}} \\  
  \caption{Calibration Process and Simulation Process}
  \label{fig:calibration}
  \scriptsize{}
\end{figure}

\section{Result} \label{sec:result}
\begin{figure}[thpb!]
\small
\centering
    \subfloat{\includegraphics[width= 0.9\linewidth]{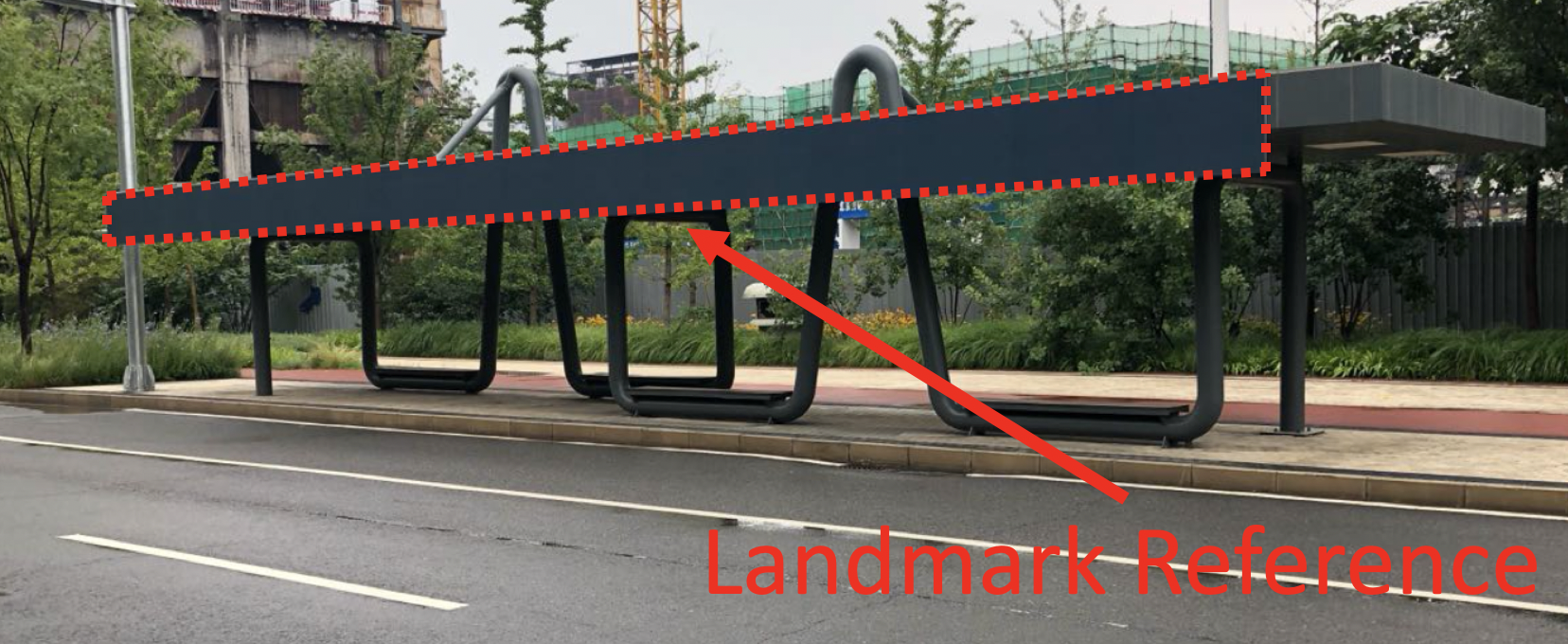}} \\    
  \caption{Landmark Example}
  \label{fig:landmark1}
  \scriptsize{}
\end{figure}

\begin{figure}[thpb!]
\small
\centering
    \subfloat{\includegraphics[width= 0.7\linewidth]{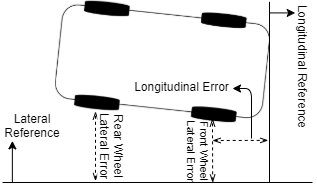}} \\    
  \caption{Measurement Method}
  \label{fig:measurement}
  \scriptsize{}
\end{figure}

\subsection{Apparatus and Testing Scenario}
For the field tests, we used an electronic vehicle with a dimension of 5m (length) * 2m (width) * 2.2m (height) with a drive-by-wire system, and an X86 computer running Apollo autonomous driving system with an architecture shown in Figure \ref{fig:workflow}, and multiple sensors that are designed for Level 4 autonomous driving systems.
The Error Feedback System uses an existing Hesai Pandar40P LiDAR \cite{Hesai40P} equipped on top of the vehicle. Other sensors are mainly used by other parts (such as the perception module) of the Apollo system \cite{Apollo2017}. 
Figure \ref{fig:landmark1} illustrates the testing scenario. As described in previous sections, this paper focuses on lateral precision (with respect to a reference). We hence tested the system through a typical bus-stop scenario. 
Note that it is a real-world, everyday bus platform built years ago, with no additional design except the grey board (the red rectangle in Figure \ref{fig:landmark1}). 
The board was added by the testing team to provide a flat and smooth surface for the LiDAR. The board was approximately 14.6 meters long, 0.6 meters wide, and in parallel with the road. During driving, the 12th ray from the top LiDAR, which is horizontal according to the Pandar40P manual, was used to measure the lateral distance between the ego vehicle and the board. To summary, we basically picked a random bus station with minimum modifications (spent 2 hours to mount the board) for the testing purpose.

\begin{table*}[]
\caption{Precision of Proposed Control Module under Standard Condition}
\label{table:normal_condition}
\centering
\small
\begin{tabular}{|c|c|c|c|c|c|}
\hline
\multirow{2}{*}{} & Longitudinal & \multicolumn{3}{c|}{Lateral Error (cm)} & \multirow{2}{*}{Heading Error} \\ \cline{3-5}
 & Error (cm) & Front Wheel & Back Wheel & LiDAR & (rad) \\ \hline
Mean & 0.2 & -0.7 & 1.7 & 0.3 & 0.0061 \\ \hline
STD & 10.1 & 0.7 & 0.9 & 0.9 & 0.0013 \\ \hline
\end{tabular}
\end{table*}

\begin{table*}[]
\caption{Precision of Human Drivers under Standard Condition}
\label{table:human_driver}
\centering
\small
\begin{tabular}{|c|c|c|c|c|c|}
\hline
\multirow{2}{*}{} & Longitudinal & \multicolumn{3}{c|}{Lateral Error (cm)} & \multirow{2}{*}{Heading Error} \\ \cline{3-5}
 & Error (cm) & Front Wheel & Back Wheel & LiDAR & (rad) \\ \hline
Mean & N/A & -0.5 & -0.3 & N/A & 0.0005 \\ \hline
STD & N/A & 4.2 & 4.0 & N/A & 0.0005 \\ \hline
\end{tabular}
\end{table*}

\begin{table*}[]
\caption{Precision of Original Control Module under Standard Condition}
\label{table:original_control}
\centering
\small
\begin{tabular}{|c|c|c|c|c|c|}
\hline
\multirow{2}{*}{} & Longitudinal & \multicolumn{3}{c|}{Lateral Error (cm)} & \multirow{2}{*}{Heading Error} \\ \cline{3-5}
 & Error (cm) & Front Wheel & Back Wheel & LiDAR & (rad) \\ \hline
Mean & N/A & -3.7 & -4.0 & N/A & -0.0007 \\ \hline
STD & N/A & 3.1 & 3.9 & N/A & 0.0072 \\ \hline
\end{tabular}
\end{table*}

\subsection{Experiment and Measurement}
As long as the ego vehicle was close to the bus stop, \textit{i.e.} LiDAR detects the board, the proposed method would be triggered (Fig. \ref{fig:workflow}). 
The system then continually adjusted its steering wheel according to the lateral and heading error feedback (see Method Section). For convenience, we used the existing (not drawn by the testing team), middle, long, white, solid lane line in Fig. \ref{fig:landmark1} as the reference for the control module. 
That means, the lateral error from LiDAR was first subtracted by a constant offset, \textit{i.e.} the lateral distance between the board and that lane line, before it was fed into the control module. 
Consequently, the ego vehicle drove towards the lane line, rather than crashing with the board. The lateral and heading errors were then measured with respect to the lane line. Fig. \ref{fig:measurement} shows how results were measured. 
We first recorded the lateral error (with respect to the lane line) from LiDAR after the ego vehicle fully stopped. We also measured the lateral error of front and rear wheels through an L-ruler (Fig.\ref{fig:measurement}), hence small human error on measurement, \textit{i.e.} $\pm$ 1cm, is expected. 
The longitudinal error was measured with respect to a horizontal line drawn by the testing team. Heading error was calculated in a way that:
\begin{equation}
Heading_{error} = \arctan (\cfrac{Lat\_error_{back} - Lat\_error_{front}}{wheelbase})
\label{eq:heading_error}
\end{equation}

Experiments were primarily carried out under standard condition, in which all of below should be satisfied: 
\begin{itemize} 
\item Half load.
\item Typical weather with no rain or snow.
\item Dry road surface.
\item Irrelevant to light condition.
\end{itemize}
\subsection{Results}
We expected lateral and heading error to be around zero with small means and standard deviations (std. for short). The lateral error measured by LiDAR was set to align with that of the rear wheel, which means those two values should match in principle. 
That said, one should consider that both measurement, \textit{i.e.} LiDAR and L-ruler, have accuracy around 1 to 2 centimeters. Forty consecutive trials were carried out with the proposed method under standard conditions. 
Table \ref{table:normal_condition} shows the mean and std. for both lateral error and heading error. Both the front wheel and rear wheel were in the zone of the target $\pm$ 5cm. 
The std. of lateral error is even less than 1cm. The lateral error measured by LiDAR is close to that by L-ruler, given the measurement accuracy. Heading error is about 0.006 rad with std. around 0.001 rad. 
The results suggest that the control module was able to provide an end-to-end lateral precision well within 3$\sigma$ $\leq$ 5cm. Interestingly, the results also imply that a steady heading error has occurred. The front-wheel was always biased to one side (in this case, right to the target) while the rear wheel biased to the opposite, leading to a 0.006 rad heading error. It is possible the performance on heading correction was limited by the capacity of localization module, and/or the system delay occurred in passing heading error from HD map, localization, and planner, to control module (see Fig $\ref{fig:workflow}$).
Another possibility, on the contrary, is that lateral error and heading error was fed into control directly, other than passing to a planning module beforehand. 
A planning module can help the control module achieve better precision. For example, a planning module can calculate a trajectory that best describes how to eliminate heading error and lateral error simultaneously upon the ego vehicle fully stops.
For this, one can refer to literature related to planning and model predicated control \cite{Pannocchia2010Disturbance, Afram2014Theory}. 

By comparison, we inspected whether human drivers can reach the same level of control precision with the same test vehicle in the same testing scenario.
Four test drivers were involved, all of whom were specially trained for this task. They all spent 2 to 3 years in autonomous driving test and around 1 year on this specific test vehicle. The drivers were provided a full 360-degree view, thanks to the cameras mounted all around the test vehicle. 
To help drivers perform at their best, we relaxed the requirement on longitudinal precision. Hence drivers only needed to focus on the lateral control, as the longitudinal target was not set for them. 
Thus, human drivers had an advantage over the proposed automated control module in this test. An overall of thirty trials has been conducted with them. Results (Table \ref{table:human_driver}) show that the lateral error (cm) is around -0.5 $\pm$ 4.2 (mean $\pm$ std.) for front wheel, and -0.3 $\pm$ 4.0 for rear wheel. The heading error (rad) is around 0.0005 $\pm$ 0.0005. 
To conclude, with intensive training and help from the 360-degree surrounding view, human drivers were able to maintain a sound accuracy but not precision. Finally, we also conducted fifteen trials using the original Apollo solution (LQR controller) for comparison. 
Table \ref{table:original_control} shows that the original control module (with original localization module) performed even worse than the human drivers, with -3.7 $\pm$ 3.1 for the front wheel, -4.0 $\pm$ 3.9 for the rear wheel, and -0.0007 $\pm$ 0.0072 for heading.

\section{Discussion and Feature Work}
Since the Error Feedback System integrates both the localization module (for heading estimation) and the LiDAR module (for lateral error), the localization module might be the root cause for the steady heading error. One solution is to use two LiDARs, one near the front wheel and one near the rear wheel, to measure the lateral error simultaneously in those two positions, from which heading  (with respect to the target) can also be estimated. 
Meanwhile, a planning module would be beneficial as it can combine multiple optimization techniques, for example, a trajectory that is optimized to eliminate heading error and lateral error as ego vehicle eventually stop. 
Finally, the calibration process contributes significantly to this study. Feature work can consider incorporating both lateral calibration (such as steering wheel offset and heading offset) and longitudinal calibration \cite{9304778} to achieve an even more reliable and precise control module. 

One key contribution of this study is that, the proposed method not only outperformed
original Apollo modules but also beat specially trained and highly experienced human test
drivers. This certainly encourages the community and industries to work together on more
fields in which autonomous driving can bring practical benefits with affordable costs and
minimum modifications to existing environments.

\section{Conclusion} \label{sec:conclusion}
In this study, we integrated both localization and LiDAR techniques to achieve a precise Error Feedback System. We also implemented a lateral calibration algorithm that is able to calibrate a vehicle's steering wheel offset and heading offset in a few seconds. A simulation was built on top of this control module to fine-tune parameters. The results show that, through combining all those techniques, the lateral precision of the control module reaches a new level.
A small lateral error (cm) around -0.7 $\pm$ 0.7 (front wheel) and 1.7 $\pm$ 0.9 (rear wheel), has been achieved. 

\bibliographystyle{unsrt}
\bibliography{Fan-bibtex}

\begin{thebibliography}{10}

\bibitem{Chen2015DeepDriving}
Chenyi Chen, Ari Seff, Alain Kornhauser, and Jianxiong Xiao.
\newblock Deepdriving: Learning affordance for direct perception in autonomous
  driving.
\newblock 2015.

\bibitem{Casser2019Unsupervised}
Vincent Casser, Soeren Pirk, Reza Mahjourian, and Anelia Angelova.
\newblock Unsupervised monocular depth and ego-motion learning with structure
  and semantics.
\newblock In {\em International Workshop on Visual Odometry Computer Vision
  Applications Based on Location Clues}, 2019.

\bibitem{Gordon2019Depth}
Ariel Gordon, Hanhan Li, Rico Jonschkowski, and Anelia Angelova.
\newblock Depth from videos in the wild: Unsupervised monocular depth learning
  from unknown cameras.
\newblock In {\em 2019 IEEE/CVF International Conference on Computer Vision
  (ICCV)}, 2019.

\bibitem{Sun2019Scalability}
Pei Sun, Henrik Kretzschmar, Xerxes Dotiwalla, Aurelien Chouard, Vijaysai
  Patnaik, Paul Tsui, James Guo, Yin Zhou, Yuning Chai, and Benjamin Caine.
\newblock Scalability in perception for autonomous driving: Waymo open dataset.
\newblock 2019.

\bibitem{2020PointContrast}
Saining Xie, Jiatao Gu, Demi Guo, Charles~R. Qi, Leonidas~J. Guibas, and
  Or~Litany.
\newblock {\em PointContrast: Unsupervised Pre-training for 3D Point Cloud
  Understanding}.
\newblock 2020.

\bibitem{Alahi2016Social}
Alexandre Alahi, Kratarth Goel, Vignesh Ramanathan, Alexandre Robicquet, and
  Silvio Savarese.
\newblock Social lstm: Human trajectory prediction in crowded spaces.
\newblock In {\em 2016 IEEE Conference on Computer Vision and Pattern
  Recognition (CVPR)}, 2016.

\bibitem{Vemula2018Social}
Anirudh Vemula, Katharina Muelling, and Jean Oh.
\newblock Social attention: Modeling attention in human crowds.
\newblock In {\em 2018 IEEE international Conference on Robotics and Automation
  (ICRA)}, 2018.

\bibitem{Deo2018How}
Nachiket Deo, Akshay Rangesh, and Mohan~M. Trivedi.
\newblock How would surround vehicles move? a unified framework for maneuver
  classification and motion prediction.
\newblock {\em IEEE Transactions on Intelligent Vehicles}, pages 129--140,
  2018.

\bibitem{Pan2019Lane}
Jiacheng Pan, Hongyi Sun, Kecheng Xu, Yifei Jiang, Xiangquan Xiao, Jiangtao Hu,
  and Jinghao Miao.
\newblock Lane attention: Predicting vehicles' moving trajectories by learning
  their attention over lanes.
\newblock {\em arXiv e-prints}, 2019.

\bibitem{Xu2020Data}
Kecheng Xu, Xiangquan Xiao, Jinghao Miao, and Qi~Luo.
\newblock Data driven prediction architecture for autonomous driving and its
  application on apollo platform.
\newblock {\em arXiv e-prints}, 2020.

\bibitem{Benekohal1988CARSIM}
R~F Benekohal and Joseph Treiterer.
\newblock {\em CARSIM: Car-following model for simulation of traffic in normal
  and stop-and-go conditions}.
\newblock 1988.

\bibitem{Dong2011Driver}
Yanchao Dong, Zhencheng Hu, Keiichi Uchimura, and Nobuki Murayama.
\newblock Driver inattention monitoring system for intelligent vehicles: A
  review.
\newblock {\em IEEE Transactions on Intelligent Transportation Systems},
  12(2):596--614, 2011.

\bibitem{Geoffrey2011Driving}
Geoffrey, Underwood, , , David, Crundall, , , Peter, and Chapman.
\newblock Driving simulator validation with hazard perception - sciencedirect.
\newblock {\em Transportation Research Part F Traffic Psychology and
  Behaviour}, 14(6):435--446, 2011.

\bibitem{CARLA2017}
Alexey Dosovitskiy, Germ{\'{a}}n Ros, Felipe Codevilla, Antonio~M. L{\'{o}}pez,
  and Vladlen Koltun.
\newblock {CARLA:} an open urban driving simulator.
\newblock {\em CoRR}, abs/1711.03938, 2017.

\bibitem{RN290}
Brian Paden, Michal Čáp, Sze~Zheng Yong, Dmitry Yershov, and Emilio Frazzoli.
\newblock A survey of motion planning and control techniques for self-driving
  urban vehicles.
\newblock {\em IEEE Transactions on Intelligent Vehicles}, 1(1):33--55, 2016.

\bibitem{9304778}
F.~{Zhu}, X.~{Xu}, L.~{Ma}, D.~{Guo}, X.~{Cui}, and Q.~{Kong}.
\newblock Autonomous driving vehicle control auto-calibration system: An
  industry-level, data-driven and learning-based vehicle longitudinal dynamic
  calibrating algorithm.
\newblock In {\em 2020 IEEE Intelligent Vehicles Symposium (IV)}, pages
  391--397, 2020.

\bibitem{Apollo2017}
https://github.com/apolloauto/apollo.

\bibitem{Fan2018}
Haoyang Fan, Fan Zhu, Changchun Liu, Liangliang Zhang, Li~Zhuang, Dong Li,
  Weicheng Zhu, Jiangtao Hu, Hongye Li, and Qi~Kong.
\newblock Baidu apollo em motion planner.
\newblock {\em arXiv preprint arXiv:1807.08048}, 2018.

\bibitem{Durrant2006Simultaneous}
Hugh Durrant-Whyte and Tim Bailey.
\newblock Simultaneous localization and mapping: Part i.
\newblock {\em IEEE Robotics Automation Magazine}, 13(2):99--110, 2006.

\bibitem{Durrantwhyte2006Simultaneous}
Hugh~F Durrantwhyte and Tim Bailey.
\newblock Simultaneous localization and mapping (slam): part ii.
\newblock {\em IEEE Robotics and Amp Amp Automation Magazine}, 13(2):99 -- 110,
  2006.

\bibitem{8461224}
G.~{Wan}, X.~{Yang}, R.~{Cai}, H.~{Li}, Y.~{Zhou}, H.~{Wang}, and S.~{Song}.
\newblock Robust and precise vehicle localization based on multi-sensor fusion
  in diverse city scenes.
\newblock In {\em 2018 IEEE International Conference on Robotics and Automation
  (ICRA)}, pages 4670--4677, 2018.

\bibitem{Narayanan2014Methods}
Kidambi Narayanan, R.~L. Harne, Fujii Yuji, Gregory~M. Pietron, and K.~W. Wang.
\newblock Methods in vehicle mass and road grade estimation.
\newblock {\em SAE International Journal of Passenger Cars - Mechanical
  Systems}, 7(3):981--991, 2014.

\bibitem{Levinson2011Towards}
Jesse Levinson, Jake Askeland, Jan Becker, Jennifer Dolson, David Held, Sören
  Kammel, J.~Zico Kolter, Dirk Langer, Oliver Pink, and Vaughan Pratt.
\newblock Towards fully autonomous driving: Systems and algorithms.
\newblock 2011.

\bibitem{Hesai40P}
https://www.hesaitech.com/en/pandar40p.

\bibitem{2019L}
Weixin Lu, Yao Zhou, Guowei Wan, Shenhua Hou, and Shiyu Song.
\newblock L3 -net: Towards learning based lidar localization for autonomous
  driving.
\newblock In {\em 2019 IEEE/CVF Conference on Computer Vision and Pattern
  Recognition (CVPR)}, 2019.

\bibitem{Zhou2020DA4AD}
Yao Zhou, Guowei Wan, Shenhua Hou, Li~Yu, Gang Wang, Xiaofei Rui, and Shiyu
  Song.
\newblock Da4ad: End-to-end deep attention-based visual localization for
  autonomous driving.
\newblock 2020.

\bibitem{LQR1974}
Huibert Kwakernaak, Raphael Sivan, and Bjor Tyreus.
\newblock Linear optimal control system.
\newblock {\em Journal of Dynamic Systems, Measurement, and Control}, 96:373,
  10 1974.

\bibitem{Pannocchia2010Disturbance}
Gabriele Pannocchia and James~B. Rawlings.
\newblock Disturbance models for offset-free mpc control.
\newblock {\em Aiche Journal}, 49(2):426--437, 2010.

\bibitem{Afram2014Theory}
Abdul Afram and Farrokh Janabi-Sharifi.
\newblock Theory and applications of hvac control systems – a review of model
  predictive control (mpc).
\newblock {\em Building and Environment}, 72(feb.):343--355, 2014.

\end{thebibliography}

\end{document}